\documentclass[letterpaper, 10 pt, conference]{ieeeconf}  

\IEEEoverridecommandlockouts                              

\usepackage{graphicx} 
\usepackage{amsmath} 
\usepackage{amssymb}  
\usepackage{algorithm,algorithmic}
\usepackage{url}
\usepackage{color}
\newcommand{\todo}[1]{}
\renewcommand{\todo}[1]{{\color{red}Todo: {#1}}}
\usepackage[export]{adjustbox}
\usepackage[noadjust]{cite}

\makeatletter
\newcommand\fs@spaceruled{\def\@fs@cfont{\bfseries}\let\@fs@capt\floatc@ruled
  \def\@fs@pre{\vspace{0.5\baselineskip}\hrule height.8pt depth0pt \kern2pt}%
  \def\@fs@post{\kern2pt\hrule\relax}%
  \def\@fs@mid{\kern2pt\hrule\kern2pt}%
  \let\@fs@iftopcapt\iftrue}
\makeatother

\DeclareMathOperator*{\argmin}{arg\,min}

\title{\LARGE \bf
In-Hand Object Pose Tracking via Contact Feedback and GPU-Accelerated Robotic Simulation
}

\author{Jacky Liang$^{1}$, Ankur Handa$^{2}$, Karl Van Wyk$^{2}$, Viktor Makoviychuk$^{2}$, Oliver Kroemer$^{3}$, Dieter Fox$^{2}$
\thanks{$^{1}$Carnegie Mellon University. Work done during internship at NVIDIA.
        {\tt\small jackyliang@cmu.edu}}%
\thanks{$^{2}$NVIDIA Research
        {\tt\small \{ahanda, vmakoviychuk, kvanwyk, dieterf\}@nvidia.com}}%
\thanks{$^{3}$Carnegie Mellon University.
        {\tt\small okroemer@andrew.cmu.edu}}%
}

\begin{document}

\setlength{\textfloatsep}{0.2cm}
\setlength{\floatsep}{0.2cm}

\maketitle
\thispagestyle{empty}
\pagestyle{empty}

\begin{abstract}
Tracking the pose of an object while it is being held and manipulated by a robot hand is difficult for vision-based methods due to significant occlusions.
Prior works have explored using contact feedback and particle filters to localize in-hand objects. However, they have mostly focused on the static grasp setting and not when the object is in motion, as doing so requires modeling of complex contact dynamics.
In this work, we propose using GPU-accelerated parallel robot simulations and derivative-free, sample-based optimizers to track in-hand object poses with contact feedback during manipulation.
We use physics simulation as the forward model for robot-object interactions, and the algorithm jointly optimizes for the state and the parameters of the simulations, so they better match with those of the real world.
Our method runs in real-time (30Hz) on a single GPU, and it achieves an average point cloud distance error of $6$mm in simulation experiments and $13$mm in the real-world ones.
View experiment videos at~\url{https://sites.google.com/view/in-hand-object-pose-tracking/}
\end{abstract}
\section{INTRODUCTION}

Performing dexterous manipulation policies benefits from a robust estimate of the pose of the object held in-hand.
Despite recent advances in pose estimation and tracking using vision feedback~\cite{xiang2017posecnn, li2018deepim, deng2019poserbpf}, in-hand object pose tracking still presents a challenge due to significant occlusions.
As such, works that require in-hand object poses are currently limited to experiments where the object is mostly visible or rely on multiple cameras~\cite{andrychowicz2018learning}, or the hand-object transform is fixed or known~\cite{rajeswaran2017learning, zhu2018dexterous}.
To mitigate the issue of visual occlusions, previous works have studied object pose estimation via contacts or tactile feedback, often by using particle filters and  knowledge of the object geometry and contact locations.
These techniques have been mostly applied to the static-grasp setting, where the object is stationary and in-grasp. 
Extending these techniques to tracking object poses during in-hand manipulation is difficult, requiring modeling of complex object-hand contact dynamics.
\begin{figure}[!ht]
    \centering
    \includegraphics[width=0.95\linewidth]{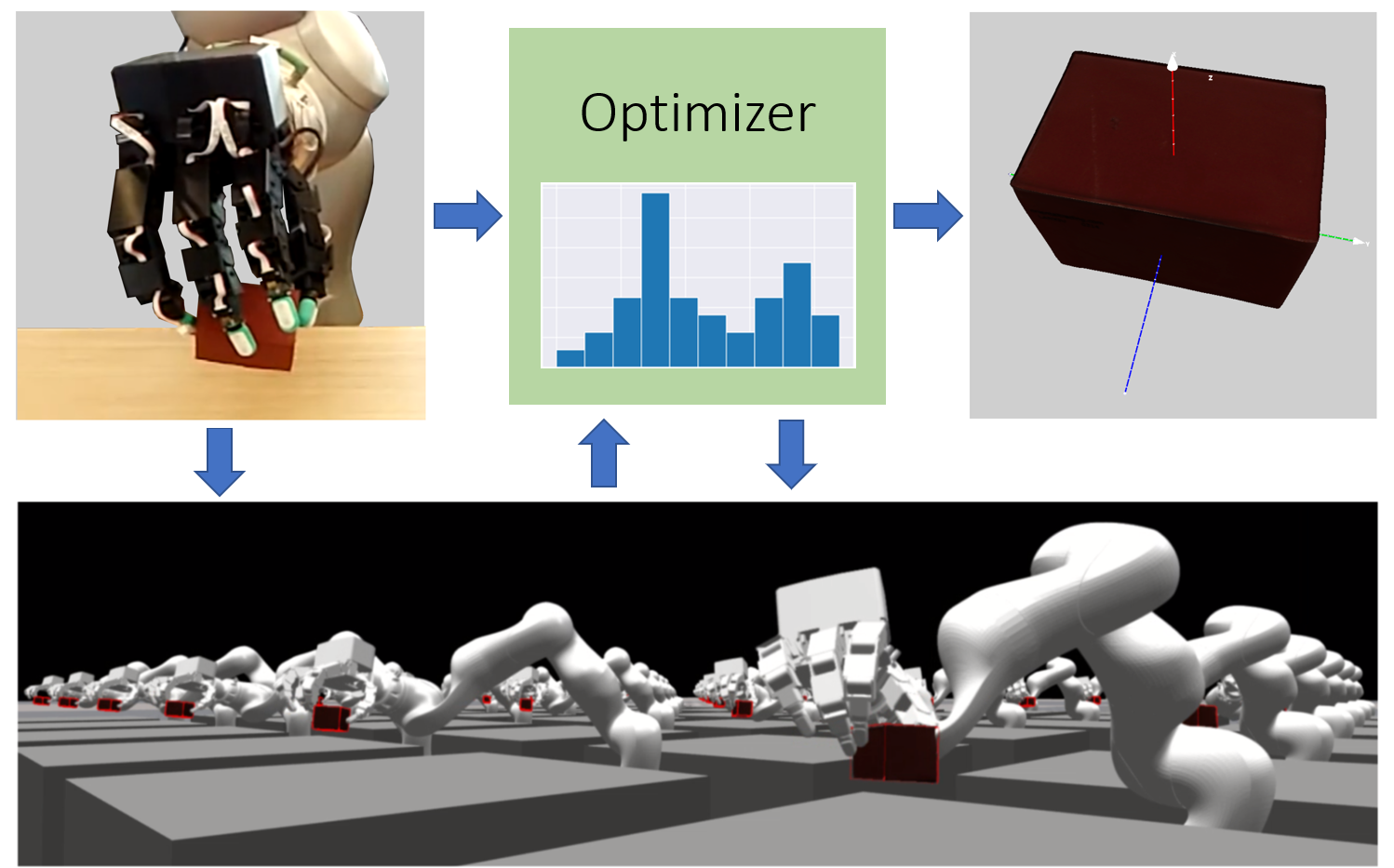}
    \caption{The proposed in-hand object pose tracking framework. Robot controls are sent to an GPU-accelerated physics simulator that runs many robot simulations in parallel, each with different physics parameters and perturbed object poses. Costs based on observations, such as contact feedback from the real world and from the simulations, are passed to a sample-based derivative-free optimizer that periodically updates the states and parameters of all simulations to better match that of the real world. At any point in time, the pose of the simulation with the lowest cost is chosen as the current object pose estimate.}
    \vspace{-5pt}
    \label{fig:teaser}
\end{figure}

To tackle the problem of in-hand object tracking during robot manipulation, we propose combining a GPU-accelerated, high-fidelity physics simulator~\cite{macklin2019non} as the forward dynamics model with a sample-based optimization framework to track object poses with contacts feedback (Figure~\ref{fig:teaser}).
First, we initialize a concurrent set of simulations with the initial states of the real robot and the initial pose of the real object, which may be obtained from a vision-based pose registration algorithm assuming the object is not in occlusion in the beginning.
The initial poses of the simulated objects are slightly perturbed and reflect the uncertainty of the vision-based pose registration algorithm.
The GPU-accelerated physics simulator runs many concurrent simulations in real-time on a single GPU.
As a given policy controls the real robot to approach, grasp, and manipulate the object in-hand, we run the same robot control commands on the simulated robots.
We collect observations of the real robot and the simulated robots, which include terms like the magnitude and direction of contacts on the robot hand's contact sensors.
Then, a sample-based optimization algorithm periodically updates the states and parameters of all simulations according to a cost function that captures how well the observations of each simulation matches with those of the real world.
In addition, the algorithm also updates simulation parameters, such as mass and friction, to further improve the simulations' dynamics models of the real world.
At any point in time, the object pose estimate is the pose of the robot-object system.

To evaluate the proposed algorithm, we collected a total of $24$ in-hand manipulation trajectories with $3$ different objects in simulation and in the real world.
For experiments, we use the Kuka IIWA7 arm with the 4-finger Wonik Robotics Allegro hand as the end-effector, with each finger outfitted with a SynTouch BioTac contact sensor.
Object manipulation trajectories are human demonstrations collected via a hand-tracking teleoperation system.
Because we have ground-truth object poses in simulation, we performed detailed ablation studies in simulation experiments to study the properties of the proposed algorithm.
For real-world experiments, we use a vision-based algorithm to obtain the object pose in the first and last frame of the collected trajectories, where the object is not in occlusion.
The pose in the first frame is used to initialize the simulations, and the pose in the last frame is used to evaluate the accuracy of the proposed contact-based algorithm.
\section{RELATED WORKS}

Prior works have studied identifying in-hand object-pose with vision only, usually by first segmenting out the robot or human hand in an image before performing pose estimation~\cite{choi2016using, kokic2019learning}.
However, vision-only approaches degrade in performance for larger occlusions.
Another approach is to use tactile feedback to aid object pose estimation.
Tactile perception can identify object properties such as materials and pose~\cite{luo2017robotic}, as well as provide feedback during object manipulation~\cite{chebotar2014learning, li2014localization, lee2019making}.

For the task of planar pushing where the object is visible, prior works have studied tracking object poses using particle filtering with contacts and vision feedback~\cite{koval2015pose}.
The authors of \cite{li2015comparative} compared a variety of dynamics models and particle filter techniques, and they found that adding noise to applied forces instead of the underlying dynamics yielded more accurate tracking results.
One work combined tactile feedback with a vision-based object tracker to track object trajectories during for planar pushing tasks~\cite{lambert2019joint}, and another applied incremental smoothing and mapping (iSAM) to combine global visual pose estimations with local contact pose readings~\cite{yu2018realtime}.

For in-hand object pose estimation with tactile feedback, many prior works have explored this problem in the ``static-grasp" context, where the robot hand grasps an object and localizes the object pose without moving.
These works can be separated into two groups: 1) using point contact locations and 2) using a full tactile map to extract local geometry information around the contacts.

To use contact location feedback for pose estimation, many methods use a variation of Bayesian or particle filtering~\cite{corcoran2010measurement, platt2011using, chalon2013online, zhang2013dynamic, javdani2013efficient, bimbo2015global, vezzani2017memory, alvarez2017tactile, ding2018hand}.
In~\cite{hebert2011fusion} the authors perform filtering jointly over visual features, hand joint positions, force-torque readings, and binary contact modes.
Similar techniques can be applied to pose estimation when the object is not held by the robot hand as well by using force probes~\cite{petrovskaya2011global, saund2017touch}.

To use tactile maps for pose estimation, earlier works used large, low-resolution tactile arrays to sense contacts in a grid~\cite{pezzementi2011object, chebotar2014learning}, while more recent works use high-resolution tactile sensors mounted on robot finger tips. 
For example, the algorithm in~\cite{bimbo2016hand} searches for similar local patches on an object surface to localize the object with respect to the contact location, and the one in~\cite{izatt2017tracking} fuses GelSight data with a point cloud perceived by a depth sensor before performing pose estimation.

By contrast to the case of static-grasps, our work tackles the more challenging problem of tracking in-hand object pose during manipulation.
Prior works have also explored this context.
In~\cite{schmidt2015depth} the authors propose an algorithm that combines contact locations with Dense Articulated Real-time Tracking (DART)~\cite{schmidt2014dart}, a depth image-based object tracking system, while in~\cite{pfanne2018fusing} the algorithm fuses contact locations with color visual features, joint positions, and force-torque readings.
The former algorithm is sensitive to initialization of the object poses, especially when the object appears small in the depth image.
The latter work conducted experiments where the objects were mostly visible, so matching visual features alone would give reasonable pose estimates.
In addition, neither work explicitly models the dynamics of the robot-object interaction, which limits the type of manipulation tasks during which the object pose can be tracked.
To address these challenges, our approach does not assume access to robust visual features during manipulation.
Instead, it uses a physics simulator to model both the kinematics and the dynamics of the robot-object system.
\section{METHOD}

\subsection{Problem Statement}
We consider the problem of tracking the pose of an object held in-hand by a robot manipulator during object manipulation.
At some time $t$, let the object pose be $p_t \in SE(3)$.
We define a physics dynamics model $s_{t+1} = f(s_t, u_t, \theta)$, where $s_t$ is the state of the world (position and velocities of rigid bodies and of joint angles in articulated bodies), $u_t\in\mathbb{R}^M$ the robot controls (we use desired joint positions as the action space), and $\theta\in\mathbb{R}^N$ the fixed parameters of the simulation (\textit{e.g.}, mass and friction).

For a simulation model $f$ that exactly matches reality given the perfect initializations of $p_0$, $s_0$, and $\theta$, pose estimation requires only playing back the sequence of actions $u_t$ applied to the robot in the simulation.
However, given our imperfect forward model and noisy pose initializations, pose estimation using our method can be improved via observation feedback.

Let $D$ be the number of joints the robot has and $L$ the number of its contact sensors.
We define the observation vector $o_t$ as the concatenation of the joint position configuration of the robot $q_t \in \mathbb{R}^D$, the position and rotation of the robot's contact sensors $P_t^{(l)}\in\mathbb{R}^{3}, R_t^{(l)}\in SO(3)$ (located on the fingertips), the force vectors of the sensed contacts $c_t^{(l)} \in \mathbb{R}^3$, the unit vector in the direction of the translational slippage on the contact surface $d_t^{(l)} \in\mathbb{R}^2$, and the binary direction of the rotational slippage on the contact surface $r_t^{(l)}\in\{0,1\}$, where $l$ indexes into the $l$th contact sensor.
The general in-hand pose estimation problem is given the current and past observations $o_{1:t}$, robot controls $u_{1:t}$, and the initial pose $p_0$, find the most probable current object pose $p_t$.

\subsection{Proposed Approach}
In this work, we leverage a GPU-accelerated physics simulator as the forward dynamics model to concurrently simulate many robot-object environments to track the in-hand object pose, and we use derivative-free, sample-based optimizers to jointly tune the state and parameters of these simulations to improve tracking performance (Algorithm~\ref{alg:alg}).
First, we obtain an estimate of the initial object pose via a vision-based object pose estimator.
We assume this pose estimator can give reliable initial pose estimate $p_0$ when the robot is not in contact with the object and when the object is not occluded, \textit{i.e.}, before grasping.
Then, given the initial object pose estimate and robot configuration, we initialize $K$ concurrent simulations, and at every timestep we copy the real robot actions $u_t$ to all $K$ simulations.
Note that the object pose can change when the hand establishes contact, and this will be modeled by the simulator.
Let the object pose and the observation of the $i$th simulation be $p_t^{(i)}$ and $o_t^{(i)}$, the ground truth observations be $o_t^\text{(gt)}$.
Given a cost function $\mathcal{C}$, we say the current best pose estimate at time $t$ is the pose of the $i^*$th simulation, $p_t^{(i^*)}$, where the $i^*$th simulation is the one that incurs the lowest average cost across some past time window $T$: 
\begin{align}
    C_i &= \frac{1}{T}\sum_{\Delta t=0}^{T-1} \mathcal{C}(o_{(t-\Delta t)}^{(i)}, o_{(t-\Delta t)}^\text{(gt)})\\
    i^* &= \argmin_{i} C_i
\end{align}
The costs are used to periodically update the simulations and their parameters.
This enables better alignment with the real robot-object system.

\subsection{Cost Function}
The desired cost function sufficiently correlates with object pose differences during in-hand object manipulation such that a lower cost corresponds to better pose estimations.
The cost function we use has the form of:
\begin{equation}
\begin{split}
    &\mathcal{C}(o_{(t-\Delta t)}^{(i)}, o_{(t-\Delta t)}^\text{(gt)})
    = w_1 ||q_t^{(i)} - q_t^\text{(gt)}||_2 \\
    &+ \sum_{l=1}^L (w_2 ||P_t^{(i, l)} - P_t^{(\text{gt}, l)}||_2 + w_3 |\Delta(R_t^{(i, l)}, R_t^{(\text{gt}, l)})|\\
    &+ w_4 (1 - \alpha_{(i, l)})+ w_5 \alpha_{(i, l)}| \Delta M(c_t^{(i, l)}, c_t^{(\text{gt}, l)})| \\
    &+ w_6 \alpha_{(i, l)}| \Delta\phi(c_t^{(i, l)}, c_t^{(\text{gt}, l)})|\\
    &+ w_7 (1 - \beta_{(i, l)}) + w_8 \beta_{(i, l)}|\Delta\phi(d_t^{(i, l)}, d_t^{(\text{gt}, l)})|\\
    &+ w_{9} (1 - \gamma_{(i, l)}) + w_{10} \gamma_{(i, l)}| r_t^{(i, l)} - r_t^{(\text{gt}, l)}|)
\end{split}
\end{equation}

For the first term in the cost function, comparing $q_t$'s between the simulated and real-world robots is useful even if they share the same $u_t$, because $q_t$ can be different depending on the collision constraints imposed by the current pose of the object in contact with the robot hand, which might make it physically impossible for a joint to reach a commanded target angle.

A contact sensor is in contact if its force magnitude is greater than a threshold.
$\alpha_{(i, l)}$ is $1$ when the binary contact state of the $l$th contact sensor of the $i$th simulation agrees with that of the real contact sensor and $0$ otherwise.
Similarly, $\beta_{(i, l)}$ is $1$ when the $l$th contact sensor of the $i$th simulation agrees with the real contact sensor in whether or not the sensor is undergoing translational slippage, $0$ otherwise; $\gamma_{(i, l)}$ is the same but for rotational slippage.

For any two vectors, $\Delta M(\cdot,\cdot)$ gives the difference of their magnitudes, and $\Delta\phi(\cdot,\cdot)$ gives the angle between them.
For any two rotations $R_a$ and $R_b$, $\Delta(R_a, R_b)$ gives the angle of the axis-angle representation of $R_a^{-1} R_b$.

The weights of the cost terms, $w_i$s, are determined such that the corresponding mean magnitude of each term is roughly normalized to $1$.

\subsection{Addressing Uncertainty and the Sim-to-Real Gap}
There are two sources of uncertainty regarding object pose estimation via simulation: 1) the initial pose estimation $p_0$ from the vision-based pose estimator is noisy and 2) there is a mismatch between the simulated and real-world dynamics, partly caused by imperfect modeling and partly caused by the unknown real-world physics parameters $\theta$.

To address the first issue of initial pose uncertainty, 1) we perturb the initial pose estimations across the different simulations by sampling from a distribution centered around the vision-based estimated pose $p_0^{(i)} \sim \mathcal{N}(p_0, \Sigma_p)$, and 2) we increase the number of simulations $K$ ($K=40$ in our experiments).
If $K$ is arbitrarily large, then it is with high probability that the true initial pose will be sufficiently represented in the set of simulations, and a well-designed cost function will select the correct simulation with the correct pose.
To perform sampling over initial object poses, we sample the translation and rotation separately.
Translation is sampled from an isotropic normal distribution, while rotation is sampled by drawing zero-mean, isotropic tangent vectors in $so(3)$ and then applying it to the mean rotation~\cite{eade2013lie}.

To address the second issue of mismatch between simulated and real-world physics (the ``sim-to-real" gap), we propose using derivative-free, sample-based optimization algorithms to tune $\theta$ during pose tracking.
Specifically, after every $T$ time steps, we pass the average costs of all simulations during this window along with the simulation state and parameters to a given optimizer.
The optimizer determines the next set of simulations with their own updated parameters.
The simulations in the next set are sampled from simulations from the current set, with some added perturbations to the simulation parameters and object pose.
Such exploration maintains the diversity of the simulations, preventing them from getting stuck in sub-optimal simulation parameters or states due to noisy observations.

Although it is desirable to have $\theta^{(i^*)}$ converge to the true $\theta^{\text{(gt)}}$, this is not necessary to achieve good pose estimation.
In addition, due to differences in simulated and real-world dynamics, we do not expect the optimal $\theta$ for reducing $\mathcal{C}$ to be their corresponding real-world values.

To optimize the parameters of the $K$ simulations and make their simulated states more closely track that of the real world, we evaluate three derivative-free, sample-based optimizers:

\subsubsection{Weighted Resampling (WRS)}
WRS forms a probability mass function (PMF) over the existing simulation states $s^{(1:K)}$ and samples $K$ times with replacement from that distribution to form the next set of simulations. 
To form the PMF, WRS applies softmax over the simulation costs:
\begin{equation}
    P(i) = \frac{\exp{-\lambda(C_i -  \min_j C_j)}}{\sum_{i=1}^K\exp{-\lambda(C_i -  \min_j C_j)}}
\end{equation}
Here, $\lambda$ is a temperature hyperparameter that determines the sharpness of the distribution.
After resampling, we perform exploration on all simulations by perturbing 1) the simulation parameters $\theta$ and 2) the object pose.

Simulation parameters are perturbed by sampling from an isotropic normal distribution around the previous parameters: $\theta^{(i)}_{\tau + 1} \sim \mathcal{N}(\theta_\tau^{(i)}, \Sigma_\theta)$, where $\Sigma_\theta$ is predefined.
The subscript $\tau$ denotes the optimizer update step (after $\tau$ update steps the simulation has ran for a total of $\tau T$ time steps).

For object pose perturbation, adding noise to the pose directly while the object is held in-hand is impractical; most delta poses would result in mesh penetration and are hence invalid.
This issue was noted in \cite{duff2011physical, li2015comparative}, and like those works we perturb the objects by applying perturbation forces to the object in each simulation, with each force sampled from a zero-mean isotropic normal distribution $v^{(i)} \sim\mathcal{N}(0, \Sigma_v)$.

\subsubsection{Relative Entropy Policy Search (REPS)}
Unlike~\cite{chebotar2018closing}, which also uses REPS~\cite{peters2010relative} to tune simulation parameters to address the sim-to-real gap, we use a sample-based variant of REPS that computes weights for each simulation and samples from a distribution formed by the softmax of those weights.
Whereas WRS uses a fixed $\lambda$ parameter to shape the distribution, REPS solves for an adaptive temperature parameter $\eta$ that best improves the performance of the overall distribution subject to $\epsilon$, a constraint on the KL-divergence between the old and updated sample distributions.

To use REPS, we reformulate the costs as rewards by setting $R_i = \max_j C_j + \min_j C_j - C_i$. 
We compute $\eta$ at every step by optimizing the dual function $g(\eta)$, and then we use $\eta$ to form the PMF:
\begin{align}
    \eta^* &= \argmin_\eta \eta\epsilon + \eta\log\frac{1}{K}\sum_{i=1}^K \exp{\frac{R_i}{\eta}} \\
    P(i) &= \frac{\exp{\frac{R_i}{\eta^*}}}{\sum_{j=1}^K\exp{\frac{R_j}{\eta^*}}}
\end{align}
After resampling, every simulation is perturbed in the same manner as in WRS.

\subsubsection{Population-Based Optimization (PBO)}
Inspired by Population-Based Training (PBT)~\cite{jaderberg2017population}, this algorithm first ranks all simulations by their average costs and finds the top $K_\text{best}$ simulations with the lowest costs.
Then, it 1) exploits by replacing the remaining $K - K_\text{best}$ simulations with copies of the $K_\text{best}$ ones, sampled with replacement, and 2) explores by perturbing the $K_\text{best}$ simulations in the same way as WRS.

PBO effectively uses a shaped cost that depends only on the relative ordering of the simulation costs and not their magnitudes, potentially making the optimizer more robust to noisy costs.

\subsection{Hyperparameters}
Each of the proposed optimizers has a distribution-shaping hyperparameter used to balance exploration with exploitation.
There are 5 additional hyperparameters for our proposed framework:
\begin{itemize}
    \item $T$ - the time steps the algorithm waits for every update.
    \item $K$ - the number of concurrent simulations.
    \item $\pmb{\theta_0}$ - the initial normal distribution over simulation parameters.
    \item $\Sigma_p$ - the diagonal covariance matrix for the normal distribution over initial pose perturbation.
    \item $\Sigma_\theta$ and $\Sigma_v$ - the diagonal covariances of normal distributions of perturbations used for exploration.
\end{itemize}
A larger $K$ is generally better than a smaller $K$, with the caveat that the resulting simulation is slower and may not be practical in application.
$\Sigma_p$ should be large enough such that the actual initial pose is well represented in the initial pose distribution. However, $K$ should be increased with a larger $\Sigma_p$ and the covariance of $\pmb{\theta_0}$ to ensure that the density of the samples is high enough to capture a wider distribution.

We note two additional trade-offs with these hyperparameters. One is the exploration-exploitation trade-off in the context of optimizing for $\theta$, and the other is the trade-off between optimizing for $\theta$ and for $p_t^{(i^*)}$.
Making $\Sigma_\theta$ or $\Sigma_v$ wider will increase the speed at which the set of simulation parameters ``move," and the optimizer will explore more than it exploits.
Increasing $T$ improves the optimization for $\theta$ as the optimizer has more samples to evaluate each simulation.
However, updating the simulation parameters too slowly might lead to drift in pose estimation if the least-cost simulation is sufficiently different from the real world, potentially leading to divergent behavior.
The worst-case divergent behavior occurs when force perturbation or some simulation parameters lead to an irrecoverable configuration, where the object falls out of the hand or brings the object into a pose such that small force perturbations cannot bring it back to the correct pose.
It is acceptable if a few samples become divergent.
Their costs will be high, so they will be discarded and replaced by ones that are not divergent during optimizer updates.

\floatstyle{spaceruled}
\restylefloat{algorithm}
\begin{algorithm}[!t]
\caption{Sample-Based Optimization for In-Hand Object Pose Tracking via Contact Feedback}
\label{alg:alg}
\begin{algorithmic}[1]
    \renewcommand{\algorithmicrequire}{\textbf{Input:}}
    \renewcommand{\algorithmicensure}{\textbf{Output:}}
    \REQUIRE $\mathcal{C}$, $T$, $K$, $s_0$, $p_0$, $\Sigma_p$, $\pmb{\theta_0}$, $\Sigma_\theta$, $\Sigma_v$, $opt$
    \STATE Initialize $K$ simulations with $s_0$, $p_0^{(i)}\sim \mathcal{N}(p_0, \Sigma_p)$, and $\theta_0\sim\pmb{\theta_0}$
    \STATE Initialize index of current best pose estimate $i^* \leftarrow 0$
    \WHILE {TRUE}
        \FOR {$t \in\{1,\hdots,T\}$}
            \STATE Obtain $u_t$ and $o_t^\text{(gt)}$ from real robot
            \FOR {$i\in\{1,\hdots,K\}$}
                \STATE Step simulation $s_{t+1}^{(i)} = f(s_t^{(i)}, u_t, \theta_\tau^{(i)})$
                \STATE Obtain simulation observation $o_t^\text{(i)}$
                \STATE Compute average cost across time window: \\
                $C_i = \frac{1}{t}\sum_{\Delta t=0}^{t-1} \mathcal{C}(o_{t-\Delta t}^{(i)}, o_{t - \Delta t}^\text{(gt)})$
            \ENDFOR
            \STATE Update index of best pose estimate:\\ $i^* \leftarrow \argmin_i C_i$
        \ENDFOR
        \STATE Update simulations according to optimizer:\\
        $s^{(1:K)}$, $\theta^{(1:K)}_{\tau+1} \leftarrow opt(s^{(1:K)}$, $\theta^{(1:K)}_\tau, C_{1:K}, \Sigma_\theta, \Sigma_v)$
    \ENDWHILE
\end{algorithmic}
\end{algorithm}

There are connections between our approach and previous works that use particle filtering.
However, prior works were mostly applied in the static-grasp setting where the forward model of the particle filter is a constant model.
Instead, we track the object during manipulation and use a physics simulator as the forward model.
In addition to tracking the object pose, the proposed algorithm also identifies the context of the forward model by tuning the simulation parameters $\theta$, which are not affected by the forward or observation models.
We focus on optimizers based on discrete samples and not continuous distributions, such as the popular Covariance Matrix Adaptation Evolution Strategy (CMA-ES)~\cite{hansen2003reducing}, because we cannot easily sample in-hand object poses from some distribution due to complicated mesh-penetration constraints imposed by contacts.
\section{EXPERIMENTS}

We evaluate the performance of our proposed approach with both simulation and the real-world experiments using an Allegro Hand mounted on a Kuka IIWA7 robot arm.
In-hand object manipulation trajectories are first collected with a hand-tracking teleoperation system, and we evaluate pose estimation errors by running our proposed algorithms offline against the collected trajectories.
These trajectories start and end with the object not in the hand and not in occlusion.
Because we have access to ground truth object poses in simulation experiments, we perform detailed ablation studies in simulation to study the effects of different hyperparameters on algorithm performance.
While we can compare pose estimation errors during the entire trajectory in simulation experiments, this is not possible in the real world.
For real-world experiments, we use PoseRBPF~\cite{deng2019poserbpf}, a recent RGB-D, particle-filter based pose estimation algorithm to obtain initial and final object poses.
We treat these initial and final object poses as ground truth and compare the final pose with the one predicted by our proposed algorithm.

We mount the 4-finger 16-DoF Allegro hand on the 7-Dof Kuka IIWA7 robot arm. 
To obtain contact feedback in the real world, we attach SynTouch BioTac sensors to each of the finger tips.
While BioTac sensors does not explicitly give force or slippage information, past works have studied how to extract such information from the sensor's raw electrode readings to predict contact force~\cite{hang2016hierarchical, sundaralingam2018robust}, slip direction~\cite{abd2018direction}, and grasp stability~\cite{su2015force, chebotar2016bigs, veiga2018hand}.
For real-world experiments, we use the trained model from~\cite{sundaralingam2018robust} to estimate force vectors $c_t$, but currently we do not estimate slippage from the BioTac sensors, so the cost function in real-world experiments do not contain the slippage terms.
Simulations were conducted on a computer with an Nvidia GTX 1080 Ti GPU, Intel i7-8700K CPU, and 16GB of memory.

\begin{figure}[!t]
\centering
    \vspace{5pt}
    \includegraphics[width=.98\linewidth]{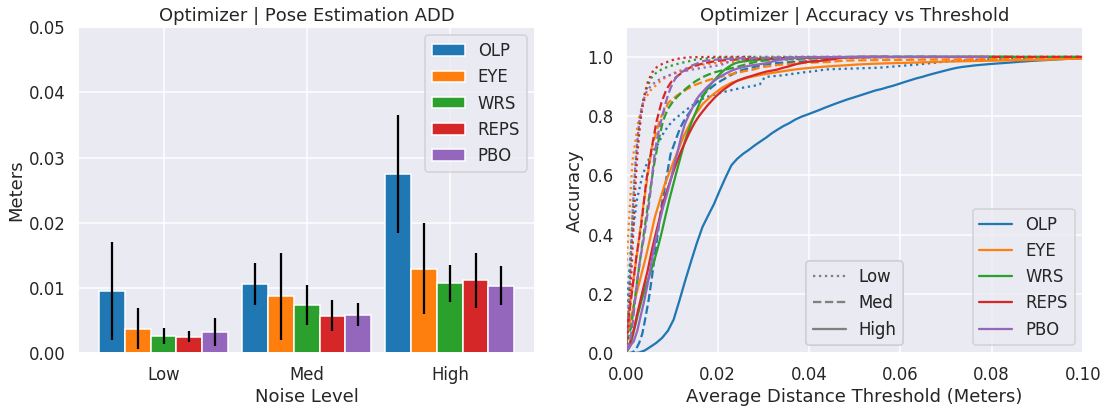}
    \vspace{-10pt}
    \caption{Pose tracking error comparison across different optimizers for initial pose noise levels in all simulation experiments. The length of the black vertical lines denote $1$ standard deviation. Optimizer methods generally have lower mean and variance of ADD, but their relative ordering varies depends on the amount of initial pose noise. REPS and PBO achieve the best ADD performance in the medium noise case - $5.8$mm and $5.9$mm respectively.}
    \vspace{-5pt}
    \label{fig:s2s_opts}
\end{figure}

We use 3 objects from the Yale-Columbia-Berkeley (YCB) objects dataset (the spam can, foam brick, and toy banana), with models, textures, and point clouds obtained from the dataset published in~\cite{xiang2017posecnn}.
These objects were chosen because they fit the size of the Allegro hand and are light enough so that robust precision grasps can be formed (we emptied the spam can to reduce its weight).

For each object, in both simulation and real-world experiments, we give 2 demonstrations of 2 types of manipulation trajectories: 1) pick and place with finger-grasp and in-hand object rotation, and 2) the same but with finger tips breaking and re-establishing contact during the grasp (finger gaiting).
This gives a total of 24 trajectories for analysis for both simulation and real-world experiments.
In both trajectory types the object undergoes translational and rotational slippage from both inertial forces and push-contacts with the table.
Each trajectory lasts about a minute. 
Given that we can run the pose estimation algorithm at about 30Hz, we obtain a total of about 2k frames per trajectory.

The teleoperation system is described in detail in a concurrent work under review.
The input to the system is a point cloud of the hand of the human demonstrator.
Then, a neural network based on PointNet++~\cite{qi2017pointnet++} maps the point cloud to an estimate of the hand's pose relative to the camera as well as the joint angles of the hand.
These estimates along with an articulated hand model~\cite{hasson19_obman} and the original point cloud are then given to DART, which performs tracking by refining upon the neural network estimates.
Finally, to perform kinematic retargetting, we solve an optimization problem that finds the Allegro hand joint angles that result in finger tip poses close to those of the human hand.

\begin{figure*}[!t]
    \vspace{2mm}
    \centerline{
    \hfill { \adjincludegraphics[width=0.235\linewidth,trim={0 0 {.5\width} 0},clip]{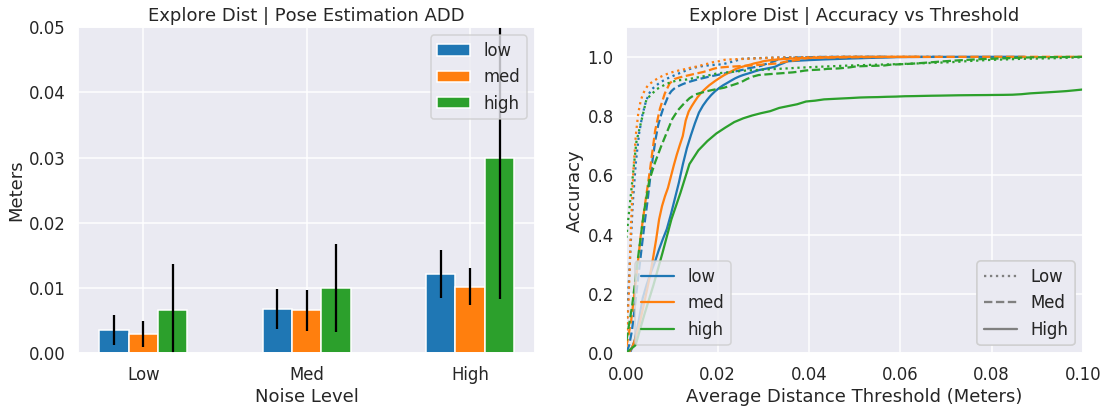} }
    \hfill { \adjincludegraphics[width=0.235\linewidth,trim={0 0 {.5\width} 0},clip]{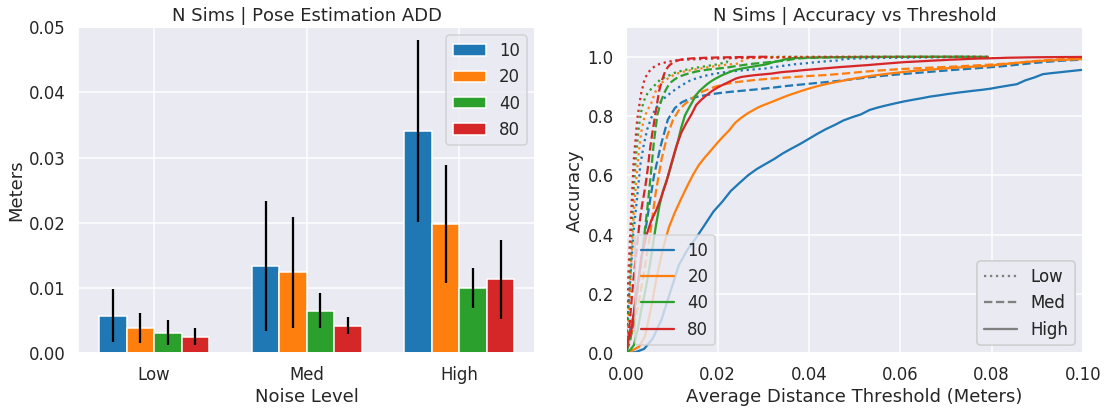} }
    \hfill { \adjincludegraphics[width=0.235\linewidth,trim={0 0 {.5\width} 0},clip]{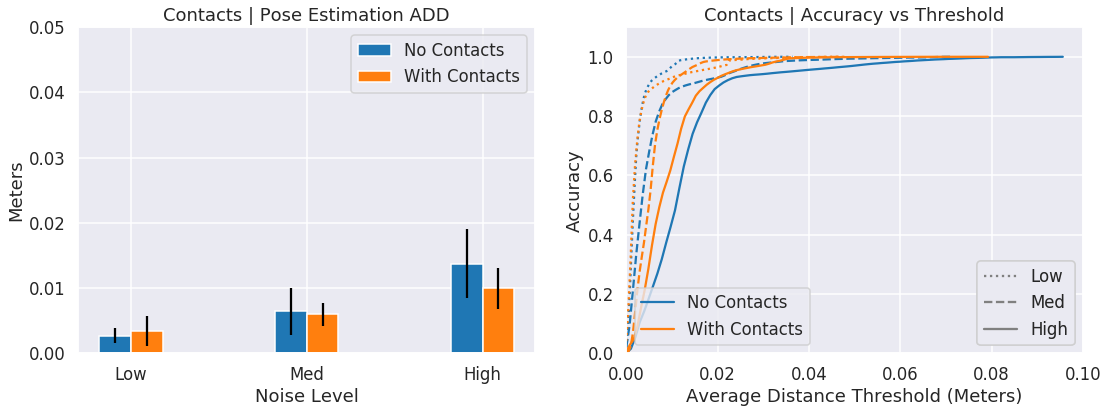} }
    \hfill { \adjincludegraphics[width=0.235\linewidth,trim={0 0 {.5\width} 0},clip]{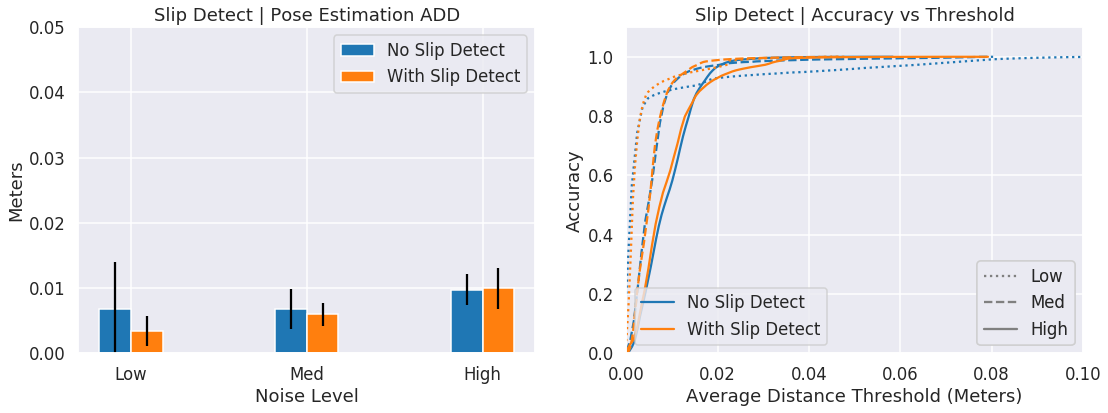} }
    \hfill
    }
    \vspace{-10pt}
    \caption{Ablation studies in simulations across different initial pose noise levels. Exploring too little or too much (how wide the $\Sigma_\theta$ and $\Sigma_v$ are.) typically result in higher mean and variance of ADD. In all other experiments we use the ``med" level of exploration. Increasing the number of simulations tend to reduce mean and variance of ADD. Contacts feedback help reduce the mean and variance of ADD, especially as noise level increases. While using slip detection helps in the case of low and medium initial pose noise, the advantage disappears when the noise is high.}
    \vspace{-15pt}
    \label{fig:s2s_ablation}
\end{figure*}

In addition to our proposed optimizers (WRS, REPS, PBO), we also evaluate the following two baselines: Open Loop (OLP) and Identity (EYE).
OLP tracks the object pose with $1$ simulation.
EYE is initialized with a set of noisy initial poses and always picks the pose of the lowest-cost simulation, but it does not perform any resampling or optimizer updates.

Similar to previous works~\cite{xiang2017posecnn, li2018deepim, deng2019poserbpf}, we use Average Distance Deviation (ADD)~\cite{hinterstoisser2012model} as the evaluation metric.
ADD computes the average distance between corresponding points in the object point cloud situated at the ground truth pose and at the predicted pose.
Unlike~\cite{xiang2017posecnn, li2018deepim, deng2019poserbpf}, we do not use its symmetric variant, ADD-S, which does not penalize pose differences across object symmetries (\textit{e.g.} for poses of a sphere that share the same translation, any rotation difference gives $0$ error).
This is desirable for resolving visual ambiguities for pose registration but not for tracking.

\subsection{Simulation Experiments}

We use Isaac Gym~\footnote{\url{https://developer.nvidia.com/isaac-gym}}, a GPU-Accelerated robotics simulator~\cite{liang2018gpu}.
The robot is controlled via a joint-angle PD controller, and we tuned the controller's gains so that the joint angle step responses are similar to those of the real robot.
To speed up simulation, we simplify the collision meshes of the robot and objects.
This is done first by applying TetWild~\cite{Hu:2018:TMW:3197517.3201353} which gives a mesh with triangles that are more equilateral, then with Quadric Edge Collapse Decimation~\cite{LocalChapterEvents:ItalChap:ItalianChapConf2008:129-136}.
Each simulation generates at most $200$ contacts during manipulation, and we run $K=40$ simulations at $30$Hz.


We performed simulation experiments with varying amounts of initial pose noise.
Three levels were tested: ``Low" has a translation standard deviation of $1$mm and a rotation standard deviation of $0.01$ radians.
``Med" is $5$mm and $0.1$ radians, and ``High" is $10$mm $1$ radian.

See Figure~\ref{fig:s2s_opts} for a comparison of the optimizers on tracking in-hand object poses across all the simulation trajectories.
ADD increases as the initial pose error increases, and the mean ADD for the optimizer-based methods tends to be lower.
While EYE sometimes achieves comparable mean ADD with the optimizer methods, the latter ones generally have much smaller error variance and max error. This result is expected as the optimizers focus the distribution of simulations towards better performing ones over time.
In the medium noise case, REPS and PBO achieve the best ADD with a mean of $5.8$mm and $5.9$mm respectively.

See Figure~\ref{fig:s2s_ablation} for results of ablation studies in simulation performed over the hyperparameters governing exploration distance (how much simulations are perturbed), the number of parallel simulations, and whether or not contact and slip detection feedback is used in the cost function.

\subsection{Real-World Experiments}

\begin{figure}[!t]
    \centering
    \includegraphics[width=\linewidth]{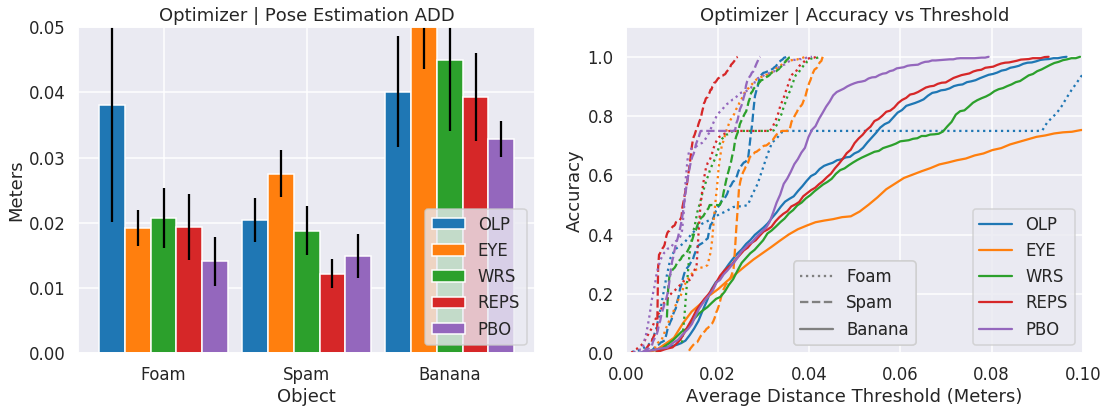}
    \vspace{-20pt}
    \caption{Real-world object pose tracking experiments across 3 objects. The adaptive optimizers generally perform better than the baselines, although all methods struggle on the banana. Due to the banana's long moment arm, precision grasps with the fingertips tend to be very unstable, and the banana undergoes significant rotational slippage or can slip out of hand in ways that are sensitive to initial conditions. The best ADD achieved with Foam is $14.1$mm by PBO, and with Spam is $12.2$mm by REPS.}
    \vspace{-5pt}
    \label{fig:r2s}
\end{figure}

We evaluate our algorithm on real-world trajectories similar to those collected in simulation.
We use PoseRBPF to register the object pose in the first and last frames of a trajectory.
The initial pose estimate is used to initialize the simulations, while the last one is used to evaluate the accuracy of our contacts-based pose tracking algorithm.
Unlike simulation experiments, the real-world experiments initialize the object pose by sampling from the distribution over object poses from PoseRBPF, so the initial pose samples correspond to the uncertainties of the vision-based pose estimation algorithm.

See Figure~\ref{fig:r2s} for real-world experiment results.
The ADDs are higher than those from simulation experiments.
This is due to both that the real-world dynamics is more dissimilar with simulations than are simulations with different parameters, and that real-world observations are noisier than those in simulations.
We observe that no optimizer is able to track the toy banana for the real-world data. 
The object's long moment arm and low friction coefficient makes its slippage behavior difficult to model precisely.
This is a failure mode of our algorithm, where if all of the simulations become divergent (\textit{e.g.} the banana rotates in the wrong direction, or falls out of hand), then the algorithm cannot recover in subsequent optimizer updates.
The best ADD achieved with Foam is $14.1$mm by PBO, and with Spam is $12.2$mm by REPS.
\section{CONCLUSION}
We introduce a sample-based optimization algorithm for tracking in-hand object poses during manipulation via contact feedback and GPU-accelerated robotic simulation.
The parallel simulations concurrently maintains many beliefs about the real world and model object pose changes that are caused by complex contact dynamics.
The optimization algorithm tunes simulation parameters during object pose tracking to further improve tracking performance.
In future work, we plan to integrate contact sensing with vision-based pose tracking in-the-loop.
\section*{ACKNOWLEDGMENT}
\scriptsize{
The authors thank Renato Gasoto, Miles Macaklin, Tony Scudiero, Jonathan Tremblay, Stan Birchfield, Qian Wan, and Mike Skolones for their help with GPU-accelerated robotic simulations, Xinke Deng and Arsalan Mousavian for PoseRBPF, and Balakumar Sundaralingam and Tucker Hermans for BioTac sensing.
This work is in part supported by the NSF Graduate Research Fellowship Program Grant No. DGE 1745016 and the Office of Naval Research Grant No. N00014-18-1-2775.
}

\clearpage
\bibliographystyle{IEEEtran}
\bibliography{citations}

\end{document}